\documentclass[sigconf]{acmart}
\usepackage{algorithm}
\usepackage[noend]{algpseudocode}

\AtBeginDocument{%
  \providecommand\BibTeX{{%
    \normalfont B\kern-0.5em{\scshape i\kern-0.25em b}\kern-0.8em\TeX}}}

\begin{document}

\setcopyright{none}
\settopmatter{printacmref=false} 
\renewcommand\footnotetextcopyrightpermission[1]{} 
\pagestyle{plain}

\title{Identifying On-road Scenarios Predictive of ADHD using Driving Simulator Time Series Data}

\author{David Grethlein}
\authornote{Both authors are student researchers.}
\email{djg329@drexel.edu}
\orcid{0002-0395-9300}
\affiliation{%
  \institution{Drexel University}
  \streetaddress{3141 Chestnut Street}
  \city{Philadelphia}
  \state{Pennsylvania}
  \country{USA}
  \postcode{19104}
}

\author{Aleksanteri Sladek}
\authornotemark[1]
\email{asladek@seas.upenn.edu}
\affiliation{%
  \institution{University of Pennsylvania}
  \streetaddress{UPenn}
  \city{Philadelphia}
  \state{Pennsylvania}
  \country{USA}
  \postcode{19104}
}

\author{Santiago Onta\~{n}\'{o}n}
\email{santi@cs.drexel.edu}
\affiliation{%
  \institution{Drexel University}
  \streetaddress{3141 Chestnut Street}
  \city{Philadelphia}
  \state{Pennsylvania}
  \country{USA}
  \postcode{19104}
}


\begin{abstract}
In this paper we introduce a novel algorithm called \emph{Iterative Section Reduction} (ISR) to automatically identify sub-intervals of spatiotemporal time series that are predictive of a target classification task. Specifically, using data collected from a driving simulator study, we identify which spatial regions (dubbed \emph{sections}) along the simulated routes tend to manifest driving behaviors that are predictive of the presence of \emph{Attention Deficit Hyperactivity Disorder} (ADHD). Identifying these sections is important for two main reasons: (1) to improve predictive accuracy of the trained models by filtering out non-predictive time series sub-intervals, and (2) to gain insights into which on-road scenarios (dubbed \emph{events}) elicit distinctly different driving behaviors from patients undergoing treatment for ADHD versus those that are not. Our experimental results show both improved performance over prior efforts ($+10\%$ accuracy) and good alignment between the predictive sections identified and scripted on-road events in the simulator (negotiating turns and curves). 
\end{abstract}





\maketitle

\section{Introduction}

The long term goal of this work is to enhance driver safety through the automatic and non-invasive identification of drivers with cognitive impairments, with a special focus on uncontrolled symptoms of {\em Attention Deficit Hyperactivity Disorder} (ADHD). Specifically, in this paper we present a new algorithm for identifying which spatial regions (we refer to them as \emph{sections}) of each simulated route are predictive of untreated forms of ADHD in adolescent drivers. These sections correspond to the types of on-road scenarios (referred to as \emph{events}) in the simulator that elicited perceivably different behaviors between the classes of drivers; useful for designing interventions. 

\emph{Motor Vehicle Collisions} (MVCs) continue to be a leading cause of death in the United States \cite{kochanek2019national}, having claimed the lives of over 36,000 people in 2018 alone \cite{US_DOT_19}. It is estimated that for each person who is killed in an MVC, there are 8 times as many who are hospitalized as result of one \cite{CDC_20}. On top of the loss of human lives, MVC-related deaths in 2013 contributed to \$44 billion in medical care and work loss costs. In order to address this, traffic safety experts have been the on-road conditions surrounding MVCs and who is at risk of being in one \cite{mcdonald2014comparison}. Armed with this knowledge, driving instructors can design meaningful interactions to mitigate risk-generating behaviors behind the wheel \cite{aduen2019expert}.

Adolescent drivers are 3 times as likely to be involved in a MVC than others \cite{iihs2019}. Those with cognitive impairments to their executive functions like ADHD are a particularly vulnerable population \cite{walshe2017executive,walshe2019working}. Some of the indicators of driving with ADHD and side-effects of taking treatment for it have been studied in clinical settings \cite{sobanski2008driving,groom2015driving}; though such methods of detection often rely on invasive techniques (eye-tracking, blood tests, or Electroencephalogram data) to render diagnoses \cite{tenev2014machine}. These approaches have made varying levels of success, typically classifying adults with more progressed forms of ADHD. In this work, we aim to non-invasively identify which adolescent drivers (age 18-24) being medicated for less-advanced forms of ADHD may require corrective instruction to avoid on-road hazards both in the simulator and in real life.

%
The immediate goal of this work is to deliver feedback to driving simulator designers detailing which of the scripted events they use effectively detect potentially dangerous drivers. We build off existing work done to design common crash scenarios as events to assess adolescent driver attention and skill levels in a driving simulator \cite{mcdonald2012using}. Our work departs from these efforts in that we intend to uncover the most discriminatory on-road scenarios without any \emph{a priori} assumptions of which will work well and which won't. 

The remainder of this document is laid out as follows. First, we review several existing tools to overcome challenging time series classification tasks, illustrating how our work relates to them. Next, we briefly describe the dataset analyzed in this study. Then we introduce the \emph{Iterative Section Reduction} (ISR) algorithm used to reduce each simulated route to a sequence of sections, where non-predictive events along the route are no longer conflated with predictive ones. We then catalog several evaluation metrics for determining the relative success of each set of ensembles formed, along with the motivations for choosing the considered metrics. After which point, we detail the grid search hyper-parameter values that define the exact manner in which ISR ensembles were formed from multiple section-specific classifiers. Subsequently, we present our best classification results for each route and relate our findings back to the types of on-road scenarios that best separated the classes of drivers examined. Finally we review the outcomes from all experiments conducted to draw our conclusions. We discuss the identified ADHD predictors, positing potential next steps for both ISR and the time series classification task at hand. 
\section{Background}
\emph{Attention Deficit Hyperactivity Disorder} (ADHD) cognitive impairment has been linked to risk-generating driving behaviors such as being easily distracted and unawareness of one's surrounding\cite{chang2014serious}. There is also evidence suggesting that pharmacological treatments have been partially beneficial in correcting these behaviors\cite{fuermaier2017driving}. Some of the most effective clinical methods for detecting ADHD, for the purpose of treatment, rely on invasive \emph{Functional Magnetic Resonance Imaging} (fMRI) technologies that require specialized equipment and trained staff to operate it \cite{iannaccone2015classifying,raviprakash2019deep}.

A \emph{time series} $T$ is an ordered sequence of measurements of some quantity (or quantities) taken over time. We refer to a \emph{sub-interval} $T_{i \to j}$ as an ordered sub-series from within $T$ that was measured from time-step $i$ to $j$. Driving simulators are a rich source of time series data for capturing the indicators of risk-generating behaviors \cite{chandrasiri2016driving,gwak2018early}, as well being a widely available technology that requires minimal supervision to perform an assessment \cite{walshe2019comparison}. Time series output recorded in a driving simulator (throttle, brake, and steering wheel inputs) has been fed into \emph{k-Nearest Neighbor} (k-NN) classifiers and detected driver inattention under cognitive load \cite{chakraborty2016automatic}. \emph{Dynamic Time Warping} (DTW) similarity-based time series clustering has also been used for modelling and differentiating different driving behaviors by analyzing vehicle velocity in various sections of road \cite{lohrer2015building}.

\emph{Learning from observation} agents trained from the same dataset we use were able to mimic the different classes of drivers in a hierarchy of on-road contexts (e.g. intersections, traffic lights, red traffic lights) with high fidelity\cite{wong2018machine}. Similarly, researchers were able to faithfully predict how \emph{learning from demonstration} agents trained from the different classes of drivers in this dataset would behave in on-road contexts by learning state-action pairs of behaviors\cite{ontanon2017learning}. The classes of drivers in this study (described in section \ref{sec:data_desc}) have been previously shown to be non-separable when globally (using the complete time series as input) comparing simulator sessions to one another \cite{grethlein2020spatially}. We hypothesize that previous classification efforts on this dataset have failed ($< 50\%$ accuracy) as the classes of drivers largely drive uniformly except in response to certain on-road scenarios [submitted 2020; in review]. These responses are likely to be spatially concentrated around the scripted on-road hazards dispersed along each simulated route. Globally comparing driver performance (using all data recorded) in the simulator may diminish a classifier's ability to leverage those predictive responses. 

\begin{figure}[t]
    \centering
    \includegraphics[width=0.3\linewidth]{"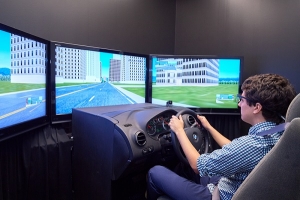"}
    \caption{Realtime Technologies RDS-1000 driving simulator used to collect the data in our study.}
    \label{fig:chop_sim}
\end{figure}

In previous experiments using this dataset, we broke down each route into a sequence of manually defined, non-overlapping sections of road [submitted 2020; in review]. We trained section-specific 1-NN and \emph{1-dimensional convolutional neural network} (1-D CNN) classifiers from the time-series sub-intervals recorded in each section in isolation from one another. We then used all section-specific classifiers trained in isolation as voters in a bagging ensemble to classify complete time series from a reserved portion of the dataset. Doing so yielded 1-NN and 1-D CNN ensembles capable of distinguishing the classes of drivers with roughly 42\% average classification accuracy across all simulated routes. The constructed classifiers also revealed high sensitivity to class labels concentrated in data recorded within sections containing turns and curves along the route. We suspect that manually defining sections cut off certain predictive behaviors. Additionally, using all sections as voters may result in sub-optimal results since the votes of classifiers trained on predictive and non-predictive sections are weighted equally.

\begin{figure*}[t]
    \centering
    \begin{tabular}{cc}
        \includegraphics[width=.35\linewidth]{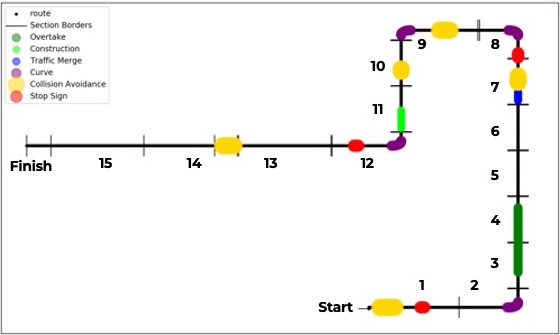} & \includegraphics[width=.35\linewidth]{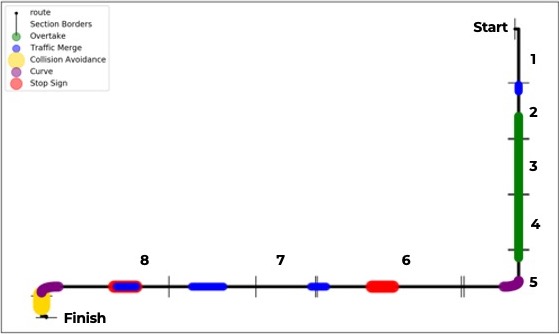} \\
        (a) Drive 1 & (b) Drive 2\\
        \includegraphics[width=.35\linewidth]{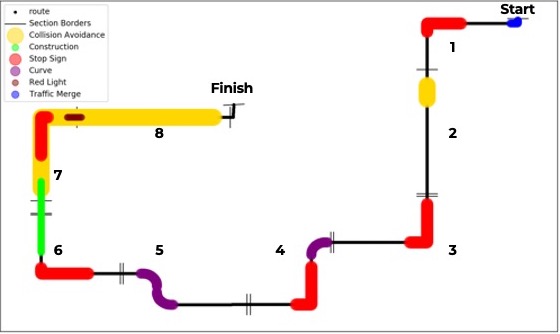} & \includegraphics[width=.35\linewidth]{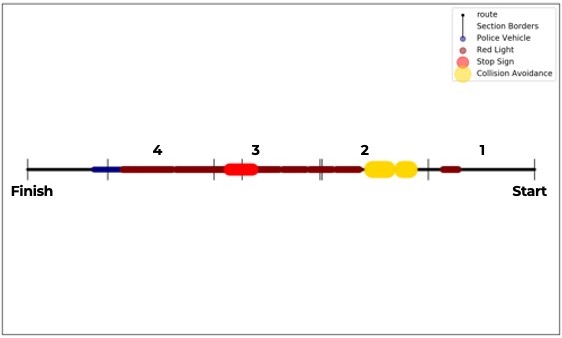} \\
        (c) Drive 3 & (d) Drive 4
    \end{tabular}
    \caption{Scripted on-road scenarios (dubbed \emph{events}) distributed along each of the 4 simulated routes. We partition the routes into 1600-meter long sections starting every 800 meters; explicitly including the last 1600 meters of the route as a section.}
    \label{fig:routes_top_level}
\end{figure*}

In a different driving simulator study, bagging ensembles composed of \emph{Long Short-Term Memory} (LSTM) neural networks have been used to successfully predict lane departures and lateral driving speed\cite{altche2017lstm}. Other \emph{recurrent neural network} (RNN) architectures have employed successfully in assessments of driving performance in simulators that also track the driver's on-screen gaze\cite{hori2016driver, manawadu2018multiclass}. Successes in these similar driver classification tasks motivated our decision to employ a deep neural network incorporating an LSTM layer in some of our experiments. Since neural networks are notorious for requiring large amounts of data to render accurate classifications and our dataset is small, we also considered several similarity based models as classifiers to act as voters in an ensemble.

\noindent
Were we to frame our analytical task as \emph{change point detection}, the prompt might read: ``there exists several modes of driving behaviors; find when and where along each route a driver transitions from one mode to the other.'' Many successful change point detection algorithms tend to monitor the \emph{log-likelihood ratio} between sequences of sub-intervals both before and after a given point in time\cite{liu2013change}. However, our task differs from these approaches in that we don't assume our drivers are changing ``modes'' or that such ``modes'' would align well with class labels. 

Framing the same task as \emph{shapelet detection} might read: ``there exists specific, commonly exhibited patterns of driving behavior that cleanly separate the classes of drivers from one another, find them.'' A \emph{shapelet} is a frequently presented pattern (very similar sub-intervals in many complete time series) that are strongly indicative of class labels for a given dataset\cite{ye2009time,mueen2011logical}. While shapelets present sophisticated insight into the composition of a labelled time series dataset, our efforts are directed at mining the predictive elements of the simulated environment itself. Moreover, our dataset is quite small ($n=30$ participants) and any commonly recurring patterns may repeat themselves several times over the route. Multiple presentations of a shapelet could make asserting any particular on-road scenario's predictive capabilities as more powerful than the others exceedingly difficult. 

Thirdly, were we to frame our analysis as the search for \emph{motifs}\cite{lin2002finding,yankov2007detecting} and \emph{discords} in the data, a prompt might read: ``identify commonly repeated patterns of behavior observed within a single driving simulator session as well as any anomalous behaviors; locate when and where they present along each route.'' \emph{Matrix profiles} are a powerful tool that find motifs and discords by relating the sub-intervals extracted via sliding window from a single time series to their most similar non-trivially matching (with chronological overlap $< 50\%$) sub-interval\cite{yeh2016matrix}. Matrix profiles can be constructed using a variety of algorithms and have event been extended to guide motif discovery using domain knowledge\cite{dau2017matrix}. We anticipate in our dataset however, largely due to prior experiments highlighting its uniformity, that all classes of drivers will share certain motifs (e.g. steering around the same number of turns and curves along a given route for all drivers). Finding \emph{discords}, or anomalous behaviors that occur in only one instance within a complete time series could hold the key to unlocking which on-road scenarios are most revealing of drivers with untreated forms of ADHD. Upon our review of the literature, we did not find any classifiers that leveraged the potential for discords to be locally concentrated within a given domain (spatially, or otherwise) as a result of common exposure to the same conditions to make predictions.

\section{Methods}

In this section, we first describe the spatiotemporal data used in our experiments. We then define our \emph{Iterative Section Reduction} (ISR) ensemble-building algorithm, which leverages a known \emph{alignment domain} (such as the spatial or temporal axis) of a time series dataset to: (1) automatically identify discriminatory sub-sections along it, and (2) use this knowledge to build an ensemble of arbitrary classifiers restricted to evaluating data from these sections. We detail several example classifiers used for composing these ensembles, and also define metrics which assess their effectiveness in identifying the classes of drivers by only examining parts of the alignment domain. Finally, we detail our grid-search strategy for tuning ISR to produce optimal results from our dataset. 

\subsection{Driving Simulator Time Series Data} \label{sec:data_desc}
 
Figure \ref{fig:chop_sim} shows the driving simulator that recorded time series data from all clinical trial participants. Our dataset was collected under National Science Foundation Grant No. 1521943 from $n=30$ participants and in its raw form consists of 91 synchronized time series channels recorded at 60 Hz . In order to reduce the overall computation time for all models considered, we reduced the dimensionality of the data via \emph{Piecewise Aggregate Approximation} (PAA)\cite{keogh2001dimensionality} to 10 Hz and only use a subset of 7 channels from each recording as input. After down-sampling, simulator sessions varied in length from 3,829 to 9,746 frames, with an average of 5,728 frames; this is equivalent to 6-17 minutes, with an average duration of 9 and a half minutes. For all experiments conducted, we used the following time series channels as some of them are frequently included when forming the basis of a simulated or naturalistic assessment of driver performance \cite{verster2011standard,choi2013effects}:
    
\begin{itemize}
    \item \emph{Throttle} - Percent depression of accelerator pedal.
    \item \emph{Brake} - Percent depression of brake pedal.
    \item \emph{Steering} - Rotation of steering wheel from resting position (in degrees).
    \item \emph{Forward Velocity} - Vehicle speed (mph).
    \item \emph{Forward Jerk} - Derivative of vehicle acceleration $(\frac{mph}{seconds^2})$.
    \item \emph{Lane Position} - Vehicle lateral offset from lane center (in feet).
    \item \emph{Heading Error} - Angle between vehicle heading vector and route following direction (in degrees).
\end{itemize}

\noindent
Our \emph{experimental} group consisted of 15 of the 30 total participants who had previously been clinically diagnosed with ADHD prior to recruitment into this study\cite{lee2018design}. These participants were recorded driving 4 planned routes (\emph{Drive 1}, ..., \emph{Drive 4}) twice; one session while receiving prescribed treatment (\emph{regulated}), and once while receiving a placebo (\emph{delayed}). The other 15 participants, referred to as the \emph{control} group, did not have confirmed ADHD diagnoses and were each recorded driving the same 4 routes once. One partial \emph{Drive 3} session from the delayed group was discarded due to simulator motion sickness, leaving a total of $N=179$ complete sessions.

Distributed along each simulated route are sequences of \emph{events} (see Figure \ref{fig:routes_top_level}), sections of the route that have been scripted to elicit responses that are telling of driver attentiveness and reactivity (collision avoidance, negotiating intersections, emergency vehicle encounters, etc.). The position of these events along the route are static, and therefore all participants driving the route encounter the same events in the same frequency and order. We view these events as potential sections to discriminate drivers with untreated forms of ADHD from those without. We compare the annotated events to the sections identified by the ISR algorithm and determine whether the events truly elicit driving behaviors predictive of ADHD. 

\begin{figure}[t]
    \centering
    \includegraphics[width=.7\linewidth]{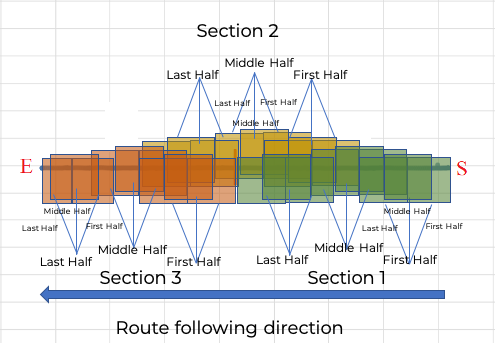}
    \includegraphics[width=.4\linewidth]{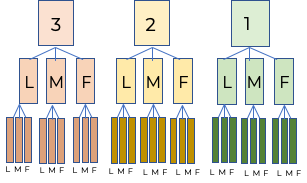}
    \caption{Planned routes can be broken down into sequences of increasingly finer-grain sections. These are then used to isolate predictive scenarios of unknown sizes distributed at unknown locations along the way.}
    \label{fig:sub_sections}
\end{figure}

To facilitate the process of iteratively dividing the simulated route into sections with ISR (see Figure \ref{fig:sub_sections}), we created a lattice data structure to represent the route. These lattices consist of ordered sequences of \emph{way-points}, linked points spaced at 5-meter intervals in the center of the road along each route. As a result, the smallest time series sub-intervals that could be extracted from our dataset will be recorded strictly within a 5-meter section of a route. For each route we define a sequence of \emph{top-level sections}, 1600-meter long sections of the route displaced with a stride of 800 meters. Top-level sections have 50\% overlap in order to give flexibility to section boundaries, since it isn't known where along the route behaviors predictive of ADHD will manifest. With 31 such top-level sections defined across all 4 routes, we expanded our dataset of 179 global time series to 1,567 clipped time series sub-intervals. 

\subsection{Iterative Section Reduction Ensembles} \label{isr}

\begin{algorithm*}[t]
\small
\caption{Iterative Section Reduction \\ \emph{X}: list of time series. \\ \emph{Y}: list of time series class labels. \\ \emph{numDev}: number of \emph{development folds} to split \emph{X} and \emph{Y} into (\emph{evaluation fold} already reserved; $numDev = k - 1$). \\ \emph{section}: data structure for clipping time series via the alignment domain. \\
\emph{depth}: current granularity of section considered, number of iterations sections have been reduced from full size. \\
\emph{maxDepth}: maximum number of iterations sections may be reduced from full size. \\
\emph{thresh}: cut-off minimum development accuracy criterion for inclusion in ensemble. \\
\emph{paradigm}: strategy for including modules in the ensemble. \\
$\theta$: Symbol representing both the type of section-specific classifiers used as modules and any relevant hyper-parameters.}
\begin{algorithmic}[1]
\Function{ISR}{X, Y, numDev, section, depth, maxDepth, thresh, paradigm, $\theta$}
\State $devMods \gets [\ ] $\Comment{Empty square brackets denote an empty list} 
\State $avgAcc \gets 0$
\State $clipX \gets clipSeriesToSection(X, section)$ \Comment{Isolate time series sub-intervals exposed to specific part of the alignment domain}
\State $SKF \gets stratifiedKFoldSplitter(numDev)$   \\ 
\For {$trainIdx, \ devIdx \in SKF.split(clipX, Y)$}\Comment{Generate $(k-1)$ independently trained classifiers for current section}
    \State $trainX, \ devX \gets clipX[trainIdx],\ clipX[devIdx]$
    \State $trainY, \ devY \gets Y[trainIdx],\ Y[devIdx]$
    \State $module \gets Classifier(\theta)$\Comment{Initialize a classifier}
    \State $module.fit(trainX,\ trainY)$ \Comment{Fit classifier to training data}
    \State $devPreds \gets module.predict(devX)$
    \State $devAcc \gets accuracy(devY,\  devPreds)$\Comment{Compute an initial estimate of performance} \\
    \If { (pardigm == ``ANY'' \textbf{ and }  devAcc $>$ thresh) \\ \indent \indent \ \ \ \ \textbf{or} paradigm == ``ALL''  } 
        \State $devMods.append(module)$ \Comment{Classifier is potential candidate for inclusion in ensemble}
        \State $avgAcc \gets avgAcc + devAcc$
    \EndIf
\EndFor \\

\If { $length(devMods) == 0$ \\ \indent \ \ \ \ \ \textbf{ or } $avgAcc/length(devMods) < thresh$ }
    \State \Return [ ]\Comment{Current section not predictive}
\EndIf    \\
\If {$depth + 1 < maxDepth$} \Comment{Max tree depth not reached, attempt to expand current section's branch}
    \State $avgAcc \gets avgAcc/length(devMods)$ 
    \State $firstHalf \gets getSectionFirstHalf(section)$
    \State $midHalf \gets getSectionMiddleHalf(section)$\Comment{Sub-divide section into 3 sub-sections (overlapping by $50\%$ of section size, stride $25\%$)}
    \State $lastHalf \gets getSectionLastHalf(section)$ \\
    \State $firstMods \gets ISR(X, Y, numDev, firstHalf, depth+1, maxDepth, avgAcc, paradigm, \theta)$
    \State $midMods \gets ISR(X, Y, numDev, midHalf, depth+1, maxDepth, avgAcc, paradigm, \theta)$\Comment{Test if sub-sections more predictive than section}
    \State $lastMods \gets ISR(X, Y, numDev, lastHalf, depth+1, maxDepth, avgAcc, paradigm, \theta)$ \\ 
    \State $subModules \gets concatenate(firstMods, midMods, lastMods)$ \\
    \If {$length(subModules) > 0$}
        \State \Return $subModules$\Comment{At least one sub-section more predictive}
    \EndIf
\EndIf \\ 
\State \Return $devModules$ \Comment{Section predictive, sub-sections not deemed more predictive}
\EndFunction
\end{algorithmic}
\label{alg:ISR}
\end{algorithm*}

\emph{Iterative Section Reduction} (ISR) is an ensemble-building algorithm (see Algorithm \ref{alg:ISR} for pseudo-code) designed to leverage an alignment domain to locate sub-intervals that are highly predictive of class labels. Its purpose is to iteratively subdivide long heterogeneous time series signals into sub-intervals. These sub-intervals are clipped using an inherent alignment domain (in our case the spatial axes), and are found by identifying the sections of the domain that elicit the most class separation. ISR constructs ensembles that are composed of multiple section-specific classifiers (dubbed as \emph{modules}) that strictly process the time series signal within their respective sections. These modules then classify sub-intervals of previously unseen data and vote (with equal weighting) which class label should be predicted for the complete simulator session.

To identify sections in the alignment domain that contain indicators of class sensitivity, ISR begins by clipping all time series in the dataset into sub-intervals using top-level sections (see \ref{sec:data_desc}). The corresponding data in each section is then partitioned (in our case by driver) into $k$ disjoint fold (ideally preserving proportions of classes in each fold). The \emph{evaluation fold} is immediately reserved from further consideration while building the ensemble, and withheld until later. The remaining $k-1$ folds are used in leave-one-out cross-validation to produce $k-1$ independently trained modules per section. Note, this means each module was trained using one of the unique combinations of $k-2$ folds of data and an initial accuracy score obtained by classifying the \emph{development fold}. This score, dubbed \emph{development accuracy}, is then saved by ISR for use as a filtering criterion for determining whether a module (and hence a specific section of the alignment domain it was trained on) will be included in the final ensemble. This score reflects the supposed discriminatory power of a module, indicating whether it could be a useful voter in an ensemble. By building multiple modules per section, we aim to mitigate any bias or stark variance that are artifacts of splitting the dataset into folds. ISR imposes a \emph{threshold} that is used to tune the degree of sufficient sensitivity to class labels needed in a section for its modules to be included in an ensemble. 

\begin{figure}[t]
    \centering
    \includegraphics[width=.7\linewidth]{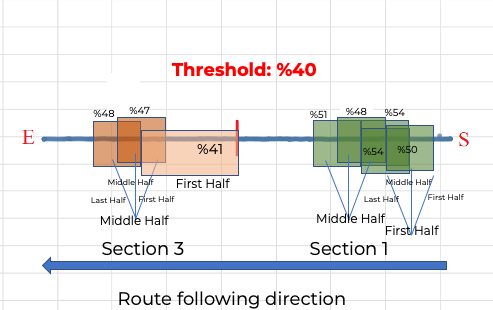}
    \includegraphics[width=.4\linewidth]{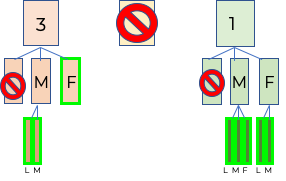}
    \caption{Predictive sub-sections extracted from the leaf nodes of top-level section trees to form an ISR ensemble.}
    \label{fig:ISR_sub_sects}
\end{figure}

If any candidate modules for inclusion are found in a section, the ISR algorithm attempts to leverage its sub-sections in an effort to further separate non-predictive parts of the alignment domain from conflating with predictive ones. Since all sections of the alignment domain are divisible into sub-sections (in our case via the lattice data structure), ISR narrows the lens of comparison for predicting class labels from time series sub-intervals. For performing the sub-division, the ISR algorithm uses a process in which a section is partitioned into three sub-sections via sliding window. Each window is 50\% the size of the original section and is collected with stride of 25\% the size of the original section (in other words, each section is divided into a \emph{first half}, \emph{middle half}, and \emph{last half}). A crucial justification for selecting this method of splitting sections into overlapping sub-sections is that it allows for flexibility in the section boundaries. This, in turn, means predictive behaviors are less likely to be split into multiple sections and more likely to captured completely within a section. 

We also experimented with two different \emph{paradigms}, strategies for determining which development fold modules are included in the ensemble: \emph{all folds per section} (ALL) and \emph{any fold above} (ANY). In the ALL paradigm, if the average development accuracy from the $k-2$ modules is above the threshold, then all are considered as candidates for inclusion in the ensemble. In the ANY paradigm, any individual fold module with development accuracy above the threshold is considered as a candidate. In both paradigms, if no potential candidates are selected, the section is deemed non-predictive and discarded from further consideration. Additionally, for both paradigms the \emph{average} development accuracy of all ensemble candidates in a section is made the new threshold value for the candidates in any subsequent sub-section to beat.

Viewing ISR from a data structures perspective, it builds a tree of potential candidate modules from sub-sections for each top-level section. The depth of these trees corresponds to a decrease in the size of the section being considered by an ensemble module. ISR then prunes branches of non-predictive sections from these trees, and harvests candidate modules from the leaf nodes remaining (see Figure~\ref{fig:ISR_sub_sects}). The algorithm is designed to leverage the fact that classifiers trained from longer time series may perform poorly, and by examining smaller sub-intervals classification accuracy can improve. Selecting a higher initial threshold value will preclude many of these smaller sub-sections from being examined, as the tree branches with modules producing insufficient development accuracy will be pruned before they are ever expanded. Choosing a lower initial threshold value will allow for the examination of more sub-section modules in the tree, and will likely allow alignment conditions to define their own lengths and predictive worth.

\subsection{Section-Specific Classifiers}

The sub-intervals detailed in section \ref{sec:data_desc} were used in two sets of experiments, where the types of modules used as voters in an ensemble were varied. In the first set of experiments, we built ISR ensembles from strictly either \emph{k-Nearest Neighbors} (k-NN) or \emph{logistic regression} (LogReg) classifiers. These modules were fed similarity matrices that relate all sub-intervals recorded in a section to one another as input. These models were selected as they require minimal parameter tuning to get up and running. In all such experiments time series sub-intervals were down-sampled using PAA to 1 Hz to further expedite similarity matrix construction. 

In the second set of experiments, we developed our own deep learning modules, dubbed \emph{DeepLSTM}. Each DeepLSTM in the ensemble is composed of the following sequence of layers: two 1-D CNN layers each with 64 filters (filter size 5, stride 1) to function as feature extractors, an average pooling layer, an LSTM layer 
(with a hidden layer size of 64) and one \emph{Fully Connected} (FC) layer of 64 neurons. The FC layer, using the Rectified Linear Unit (ReLU) activation function, feeds into an output layer of three neurons using soft-max activation, which produces the final class-wise probabilities. For all networks we used the \emph{Adam} optimizer 
and the \emph{categorical cross entropy loss function} during the training phase. To reduce overfitting, 20\% dropout\cite{dropout} layers were added between the LSTM and FC layer, as well as between the FC and output layer. Each network was fed inputs in mini-batches of 32 windows at a time, trained for 50 epochs with a learning rate of $10^{-4}$. This fixed number of training epochs was chosen given the small size of datasets fed to each DeepLSTM module and the large volume of models trained with ISR, as this reduces overfitting and computation time.

The DeepLSTM classifiers, in contrast to LogReg and k-NN, were fed the section-specific 10 Hz clippings directly. Training instances were extracted from the clippings with the sliding window method, with a window size of 30 frames (roughly 3 seconds) and a window stride of 15 frames. A small window size was chosen as many of the smallest sections considered had as little as 10 seconds worth of data for a given participant. Presented with the need to evaluate a small number of already very short time series clippings, DeepLSTM was fed the 10 Hz data to preserve both the number of overall data points, and the scarce features resulting from flattening via PAA.

{\color{blue}
\begin{figure}[t]
    \centering
    \begin{tabular}{c}
        \includegraphics[width=\linewidth]{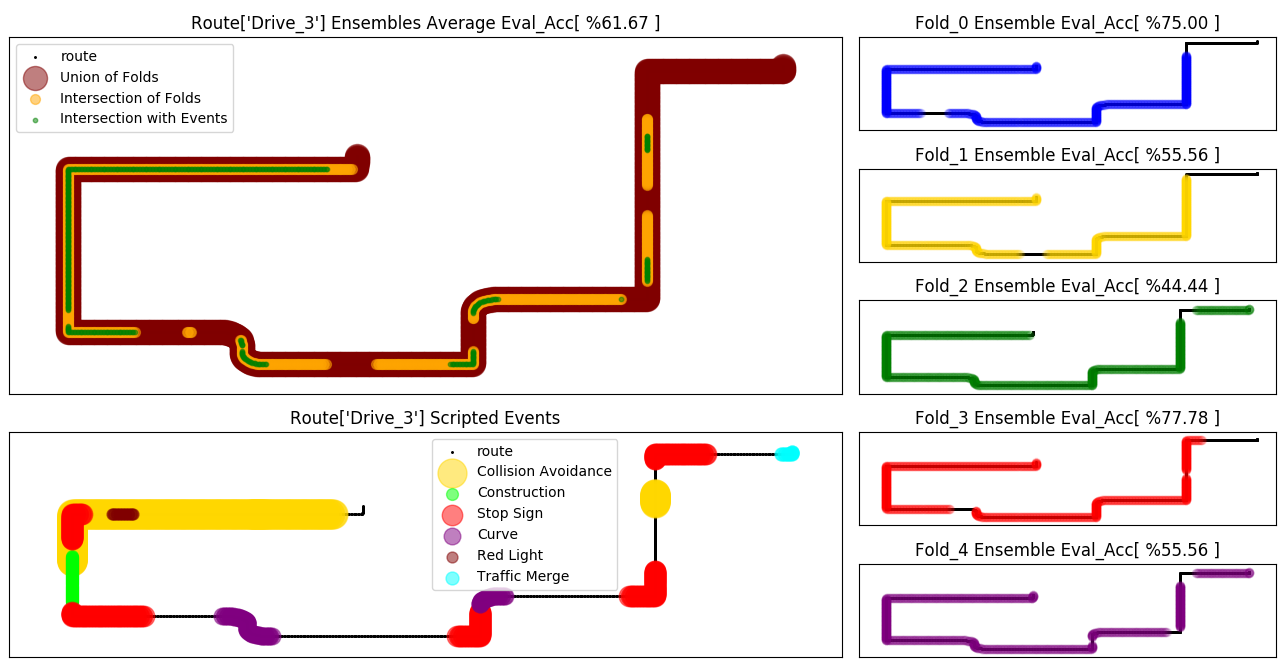} \\
        \includegraphics[width=\linewidth]{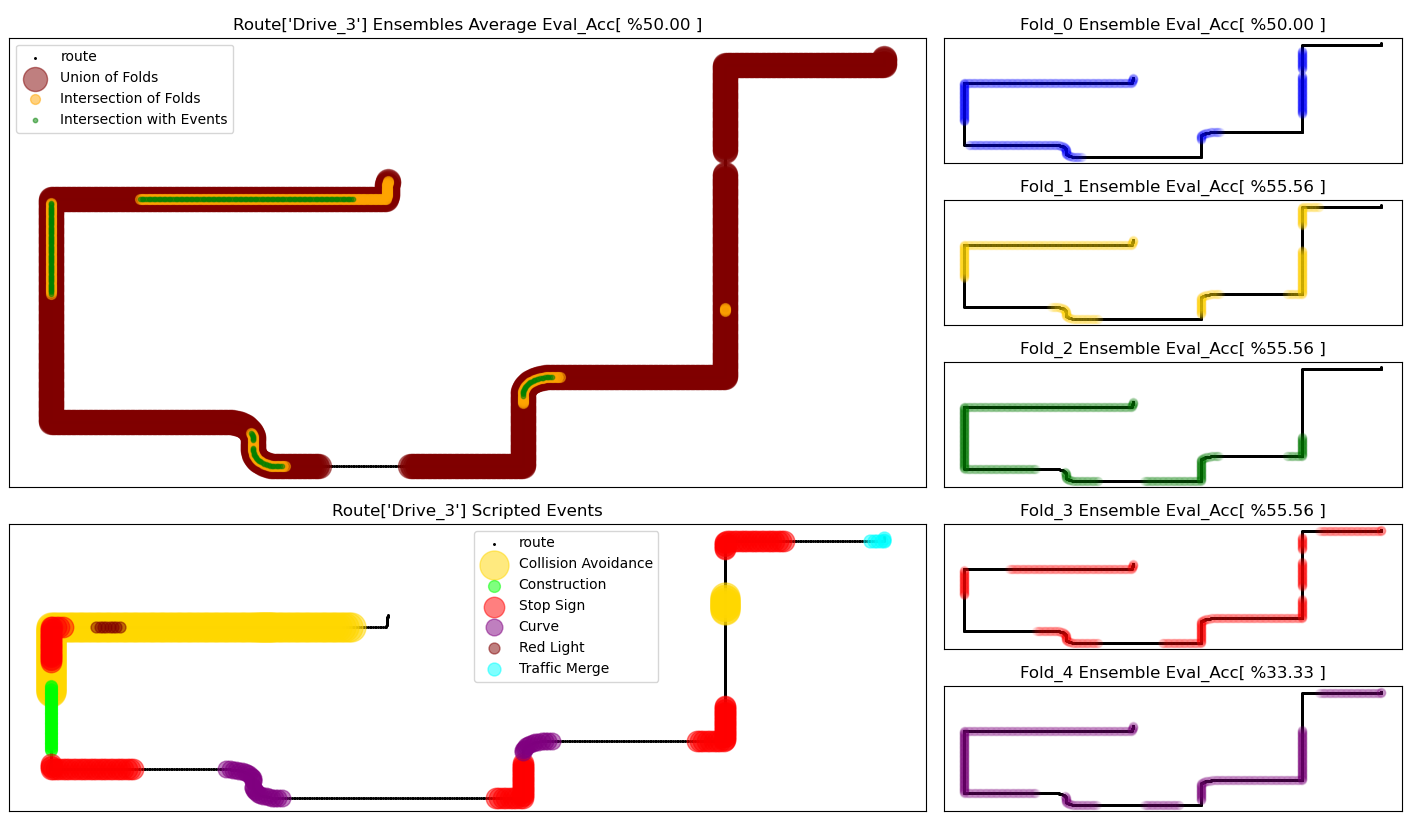}
    \end{tabular}
    
    \caption{(Top) The best performing \emph{Drive 3} ISR ensembles were built using \emph{FastDTW} with $radius=1$ as the underlying similarity function comparing time series samples, 1-NN modules, the ANY inclusion paradigm, a max depth of 3, and an initial threshold of 30\%. (Bottom) The best performing \emph{Drive 3} DeepLSTM ensembles were built using the ALL paradigm, a max depth of 3, and an initial threshold accuracy of 30\%. All 5 evaluation folds produce ensembles with high sensitivity to class labels and their intersection aligns well with scripted on-road events. This suggests that intervention designers were correct to hypothesize ADHD would noticeably affect driving behaviors in when negotiating curves, navigating a construction zone, or avoiding swerving vehicles.}
    \label{fig:fast_dtw_best}
\end{figure}
}

\subsection{Evaluation Metrics} \label{metrics}

To evaluate the performance of ISR, we added a further layer of cross-validation into ISR. For each route, we constructed $k=5$ ensembles by choosing each of the $k$ folds created by ISR once to be reserved as an \emph{evaluation fold} for a given ensemble (excluding it from the ensemble building process). After being built with the remaining $k-1$ folds, the ISR ensemble classifies previously unseen sub-intervals in the evaluation fold. Finally, ISR aggregates the votes of all member modules and produces a final classification for each complete driving simulator session examined. The reliability of these predictions is then measured in \emph{Average Evaluation Accuracy} (Acc) across the performance of all $k=5$ folds.

We also evaluated ISR using other metrics. Every ensemble built by ISR also produced a subset $S_i$ of the route lattice way-points $L$, which defines sections of the planned route found to elicit driving behaviors predictive of the driver's class. This was used to evaluate the \emph{Percent union of route} (PUoR) metric, which is the degree to which all 5 ensembles make use of the entire route (see equation \ref{eq:puor}).

\begin{equation} \label{eq:puor}
PUoR = \frac{|\bigcup_{i=1}^{k}S_i|}{|L|}
\end{equation}

We also wish to evaluate the how well do ensembles agree upon predictive scenarios across all evaluation folds. \emph{Percent Intersection with Route} (PIoR) measures the total fraction of the simulated route all evaluation fold ensembles have unanimously asserted show sensitivity to class labels (see equation \ref{eq:pior}).

\begin{equation} \label{eq:pior}
PIoR = \frac{|\bigcap_{i=1}^{k}S_i|}{|L|}
\end{equation}

We also measure the degree to which the sections selected across evaluation folds agree with one another. The \emph{Jaccard similarity of the Intersection with the Union} (JIwU) is the ratio of PIoR to PUoR and measures the consistency to which sections were deemed predictive across experiments. As each route contained the pre-determined \emph{events} depicted in Figure \ref{fig:routes_top_level}, the subset $E$ of the lattice way-points covering these events was known to us. This allowed us to analyze the overlap of ISR's unanimously-selected route sections with on-road events using \emph{Jaccard similarity of Intersection with Events} (JIwE) (equation \ref{eq:jiwe}). JIwE is defined as the fraction of lattice way-points shared by all 5 evaluation fold ensembles that overlap with the lattice way-points that encompass events (see Figure~\ref{fig:fast_dtw_best}).

\begin{equation} \label{eq:jiwe}
JIwE = \frac{|\bigcap_{i=1}^{k}S_i \cap E|}{|\bigcup_{i=1}^{k}S_i \cup E|}
\end{equation}

JIwE was motivated to quantify the degree to which the events designed by the route developers to elicit ADHD-predictive behaviors were effective in doing so. By evaluating which sections of the route all 5 ensembles ISR found to contain predictive behaviors, we can provide route designers feedback on how commonly their scripted events made a difference in separating classes of drivers. 

\begin{table*}[t]
    \centering
    \begin{tabular}{|c||l*{4}{|c}||l*{6}{|c}|}
        \hline
        Route & Similarity & Module & Paradigm & MD & Threshold & Acc & Prior Acc & PUoR & PIoR & JIwU & JIwE \\
        \hline 
        \hline
        Drive 1 & FastDTW $R=1$ & LogReg & ANY & 2 &  0 \% & \textbf{57.78 \%}  & 48.89 \% & 100 \% & 96.84 \%  & 96.84 \% & 80.45 \%  \\
        Drive 2 & SC DTW $R=15$ & LogReg & ALL & 2 &  0 \% &  44.44 \% & \textbf{51.11 \%} & 100  \% &  88.10 \% &  88.10 \% & 82.89 \%  \\
        Drive 3 & FastDTW $R=1$ & 1-NN  & ANY & 3 &  30 \% &  \textbf{61.67 \%} & 59.43 \% &  100 \% &  69.21 \% &  69.21 \% & 71.84 \% \\
        Drive 4 & DTW & 1-NN  & ANY & 4 & 35 \% &  \textbf{44.44 \%} & 37.78 \% & 100 \% &  23.82 \% &  23.82 \% & 28.37 \% \\
        \hline
    \end{tabular}
    \caption{Ensembles achieving highest 5-fold average evaluation accuracy per route from all similarity-based grid searches. \emph{Prior Acc} is the highest 5-fold classification accuracy obtained for each route so far [submitted 2020; in review]. Best classification results for each route highlighted in bold.}
    \label{tab:sim_func_best_results}
\end{table*}

\begin{table*}[t]
    \centering
    \begin{tabular}{|c||l*{2}{|c}||l*{5}{|c}|}
        \hline
        Route & Paradigm & MD & Threshold & Acc & PUoR &  PIoR & JIwU & JIwE \\
        \hline 
        \hline
        Drive 1 & ALL & 2 &  15 \% &  48.89 \% &  100 \% &  75.93 \% &  75.93 \% &  62.28 \% \\
        Drive 2 & ANY & 3 & 15 \% & 48.89 \% & 100 \% & 73.82 \% & 73.82 \% & 79.02 \% \\
        Drive 3 & ALL & 4 & 30 \% & 50.00 \% & 94.37 \% & 25.69 \% & 27.22 \% & 23.87 \% \\
        Drive 4 & ANY & 2 & 30 \% & 42.22 \% & 100 \% & 68.68 \% & 68.68 \% & 59.73 \% \\
        \hline
    \end{tabular}
    \caption{Ensembles achieving highest 5-fold average evaluation accuracy per route from all \emph{DeepLSTM} grid searches.}
    \label{tab:deep_lstm_best_results}
\end{table*}

\subsection{Hyper-parameter Grid Search}

In order to gauge the maximum sensitivity to class labels achievable, we tested hundreds of ISR parameterizations for each route and only list the ones with highest Acc in section \ref{results}. In the first set of experiments, we constructed similarity matrices to compare the driving performance of all participants to one another within a defined section. We elected to use brute force DTW and its approximations \emph{Sakoe-Chiba Dynamic Time Warping} (SC DTW) and \emph{FastDTW} to gauge how precisely time series similarity must be defined between samples to render accurate classifications \cite{sakoe1974computer,salvador2007toward}. The \emph{radius} parameter in both extensions to the DTW algorithm characterize the size of a search area for finding the alignment between time series frames that nearly minimizes the sum of residual differences, supposedly at a fraction of the computational cost. SC DTW accomplishes this by defining a band of potential frame alignments to search across the diagonal of the DTW warping cost matrix. FastDTW uses a system of reduction, projection, and refinement at multiple resolutions to define a search space of potential frame alignments with width twice the radius parameter.

For all similarity-based experiments, we permuted the 15 similarity functions (DTW, SC DTW and FastDTW both with $radius \in \{1,5,10,15,20,25,30\}$) and the types of base classifier (1-NN and LogReg). For both sets of experiments, we also permuted the \emph{Maximum Depth} (MD), i.e the number of times a section was sub-divided, of top-level section trees allowed using values 1, 2, 3, and 4. Additionally, we observed the effects of setting a higher initial threshold for the ISR algorithm with both ANY and ALL paradigms by testing the values 0\%, 5\%, 10\%, 15\%, 20\%, 25\%, 30\%, 35\%, 40\%, 45\%, and 50\%. This resulted in 2,640 unique ISR parameterizations being evaluated per route in the similarity-based experiments and 88 unique DeepLSTM parameterizations per route. We performed these grid searches and identified the initializing ISR hyper-parameters that yielded the highest Acc across all $k=5$ folds.

\section{Results} \label{results}

As a whole, ISR ensembles were only able to out-perform (average evaluation accuracy $> 44\%$ [submitted 2020; in review]) previous efforts to detect on-road scenarios highly predictive of ADHD in routes \emph{Drive 1}, \emph{Drive 2}, and \emph{Drive 3}. \emph{Drive 4} is the shortest route by distance, contains the fewest top-level sections, and doesn't have any turns or curves in the road. There wasn't any apparent trend in the degree of impact different types of modules trained or paradigms of inclusion had on evaluation accuracy. There were ISR parameterizations yielding at least partial sensitivity to ADHD class labels for all routes considered.

Table~\ref{tab:sim_func_best_results} shows the most accurate ensemble results and hyper-parameters per route from all similarity-based experiments. Table~\ref{tab:deep_lstm_best_results} shows the same but for the DeepLSTM experiments. Similarity-based ensembles tended to be more accurate for \emph{Drive 1} and \emph{Drive 3}; although DeepLSTM ensembles were the only ones able to consistently detect on-road scenarios along \emph{Drive 2} that are predictive of untreated ADHD. Overall, the \emph{control} group was easily identified by most ensembles, however the \emph{delayed} were confused with \emph{regulated} and the \emph{regulated} with the \emph{control} group. This could suggest that the medication being taken by some participants was partially effective in mitigating ADHD-associated risk-generating behaviors. 

All ensembles benefited from using at least one level of sub-sections to extract predictive scenarios. Choosing high initial threshold values yielded less accurate ISR ensembles overall, with results typically worsening significantly for threshold values exceeding the class \emph{a priori probability} of 33 \%. Being too scrutinizing of candidates for inclusion at shallow levels of the ISR tree tended to preclude the discovery of more predictive candidates at deeper levels. In the DeepLSTM experiments, less of the route was often required to classify time series samples than used in the similarity-based approaches. This meant that DeepLSTM ensembles tended to agree less frequently on the sections deemed predictive. There were no discernible trends as per which parameterizations resulted in ensembles most frequently aligning predictive sections with scripted events. However, scenarios involving negotiating stop sign intersections, curves, and avoiding collisions with other vehicles were selected as predictive of untreated ADHD in most ensembles with high classification accuracy.

\section{Discussion}

Since we only assessed one small dataset of $n=30$ drivers, our findings about which on-road scenarios elicit responses indicative of ADHD while statistically robust, must be treated as anecdotal. That being said, the ISR algorithm demonstrated a capacity for isolating predictive sub-sections of the spatial alignment domain to improve classification accuracy. Simultaneously, ISR isolated on-road events that adolescents driving with untreated forms of ADHD are likely to react to in a noticeably different manner than their peers. This is valuable information for researchers designing driving simulator routes that more effectively elicit responses from users. By only including events that have been found to be useful for driver classification tasks, both time and resources can be saved in the driving simulator design and usage processes. While ISR was shown to out-perform prior efforts at detecting ADHD in drivers, it remains unproven in other domains and against similar time series data mining techniques such as matrix profiles. 

We posit that similarity-based methods of developing ensembles performed marginally better in most cases for this small dataset as deep NNs are notoriously greedy for large sample sizes to perform well. On the other hand, we anticipate that given sufficient processing resources DeepLSTM ensembles would scale better to accommodate a larger time series dataset than the similarity-based methods, due to the exponentially increasing computation costs for similarity matrices. We also acknowledge that improved ADHD classification accuracy could likely be achieved if we had collected eye-tracking or other physiological time series channels from participants along with their driving performance data. The purpose of our study, and perhaps a limitation of it was to differentiate classes of drivers using non-invasive means.

\section{Conclusions}

First, we provided evidence that ADHD can be detected from driving behaviors without invasive technologies (brain scans, blood tests, eye tracking) requiring clinical staff supervision. Furthermore, we identified specific on-road scenarios to include as scripted events that are useful for identifying adolescent drivers with ADHD who are particularly susceptible to risk-generating behaviors behind the wheel. Second, we confirmed an independent claim that FastDTW can in actuality be slower than DTW in certain cases (see Appendix). 

We intend to employ ISR in other spatially motivated driver identification tasks, submitting that ISR may be capable of generating insights relevant to any spatial alignment domain. Examining ISR in non-spatially motivated alignment domains is necessary to determine its utility beyond our dataset.  We theorize that studying temporal alignment domains (e.g. stock prices recorded during certain hours of day trading could be predictive for determining the end of day value) could prove a valuable avenue of research for applying ISR.

\begin{acks}
Thank you to our families, friends, colleagues, and all front-line workers for sustaining us through this tough year.
\end{acks}

\bibliographystyle{ACM-Reference-Format}
\bibliography{main}


\begin{thebibliography}{42}


\ifx \showCODEN    \undefined \def \showCODEN     #1{\unskip}     \fi
\ifx \showDOI      \undefined \def \showDOI       #1{#1}\fi
\ifx \showISBNx    \undefined \def \showISBNx     #1{\unskip}     \fi
\ifx \showISBNxiii \undefined \def \showISBNxiii  #1{\unskip}     \fi
\ifx \showISSN     \undefined \def \showISSN      #1{\unskip}     \fi
\ifx \showLCCN     \undefined \def \showLCCN      #1{\unskip}     \fi
\ifx \shownote     \undefined \def \shownote      #1{#1}          \fi
\ifx \showarticletitle \undefined \def \showarticletitle #1{#1}   \fi
\ifx \showURL      \undefined \def \showURL       {\relax}        \fi
\providecommand\bibfield[2]{#2}
\providecommand\bibinfo[2]{#2}
\providecommand\natexlab[1]{#1}
\providecommand\showeprint[2][]{arXiv:#2}

\bibitem[\protect\citeauthoryear{Aduen, Cox, Fabiano, Garner, and Kofler}{Aduen
  et~al\mbox{.}}{2019}]%
        {aduen2019expert}
\bibfield{author}{\bibinfo{person}{Paula~A Aduen}, \bibinfo{person}{Daniel~J
  Cox}, \bibinfo{person}{Gregory~A Fabiano}, \bibinfo{person}{Annie~A Garner},
  {and} \bibinfo{person}{Michael~J Kofler}.} \bibinfo{year}{2019}\natexlab{}.
\newblock \showarticletitle{Expert recommendations for improving driving safety
  for teens and adult drivers with ADHD}.
\newblock \bibinfo{journal}{\emph{The ADHD report}} \bibinfo{volume}{27},
  \bibinfo{number}{4} (\bibinfo{year}{2019}), \bibinfo{pages}{8--14}.
\newblock


\bibitem[\protect\citeauthoryear{Altch{\'e} and de~La~Fortelle}{Altch{\'e} and
  de~La~Fortelle}{2017}]%
        {altche2017lstm}
\bibfield{author}{\bibinfo{person}{Florent Altch{\'e}} {and}
  \bibinfo{person}{Arnaud de La~Fortelle}.} \bibinfo{year}{2017}\natexlab{}.
\newblock \showarticletitle{An LSTM network for highway trajectory prediction}.
  In \bibinfo{booktitle}{\emph{2017 IEEE 20th International Conference on
  Intelligent Transportation Systems (ITSC)}}. IEEE, \bibinfo{pages}{353--359}.
\newblock


\bibitem[\protect\citeauthoryear{CDC}{CDC}{2020}]%
        {CDC_20}
\bibfield{author}{\bibinfo{person}{CDC}.} \bibinfo{year}{2020}\natexlab{}.
\newblock \bibinfo{title}{The Full Impact of Motor Vehicle Crashes}.
\newblock
\newblock
\urldef\tempurl%
\url{https://www.cdc.gov/motorvehiclesafety/index.html}
\showURL{%
\tempurl}


\bibitem[\protect\citeauthoryear{Chakraborty and Nakano}{Chakraborty and
  Nakano}{2016}]%
        {chakraborty2016automatic}
\bibfield{author}{\bibinfo{person}{Basabi Chakraborty} {and}
  \bibinfo{person}{Kotaro Nakano}.} \bibinfo{year}{2016}\natexlab{}.
\newblock \showarticletitle{Automatic detection of driver's awareness with
  cognitive task from driving behavior}. In \bibinfo{booktitle}{\emph{2016 IEEE
  International Conference on Systems, Man, and Cybernetics (SMC)}}. IEEE,
  \bibinfo{pages}{003630--003633}.
\newblock


\bibitem[\protect\citeauthoryear{Chandrasiri, Nawa, and Ishii}{Chandrasiri
  et~al\mbox{.}}{2016}]%
        {chandrasiri2016driving}
\bibfield{author}{\bibinfo{person}{Naiwala~P Chandrasiri},
  \bibinfo{person}{Kazunari Nawa}, {and} \bibinfo{person}{Akira Ishii}.}
  \bibinfo{year}{2016}\natexlab{}.
\newblock \showarticletitle{Driving skill classification in curve driving
  scenes using machine learning}.
\newblock \bibinfo{journal}{\emph{Journal of Modern Transportation}}
  \bibinfo{volume}{24}, \bibinfo{number}{3} (\bibinfo{year}{2016}),
  \bibinfo{pages}{196--206}.
\newblock


\bibitem[\protect\citeauthoryear{Chang, Lichtenstein, D’Onofrio,
  Sj{\"o}lander, and Larsson}{Chang et~al\mbox{.}}{2014}]%
        {chang2014serious}
\bibfield{author}{\bibinfo{person}{Zheng Chang}, \bibinfo{person}{Paul
  Lichtenstein}, \bibinfo{person}{Brian~M D’Onofrio}, \bibinfo{person}{Arvid
  Sj{\"o}lander}, {and} \bibinfo{person}{Henrik Larsson}.}
  \bibinfo{year}{2014}\natexlab{}.
\newblock \showarticletitle{Serious transport accidents in adults with
  attention-deficit/hyperactivity disorder and the effect of medication: a
  population-based study}.
\newblock \bibinfo{journal}{\emph{JAMA psychiatry}} \bibinfo{volume}{71},
  \bibinfo{number}{3} (\bibinfo{year}{2014}), \bibinfo{pages}{319--325}.
\newblock


\bibitem[\protect\citeauthoryear{Choi, Kim, Kang, Choi, Kim, Hong, Yu, Lim,
  Min, Tack, et~al\mbox{.}}{Choi et~al\mbox{.}}{2013}]%
        {choi2013effects}
\bibfield{author}{\bibinfo{person}{Jin-Seung Choi}, \bibinfo{person}{Han-Soo
  Kim}, \bibinfo{person}{Dong-Won Kang}, \bibinfo{person}{Mi-Hyun Choi},
  \bibinfo{person}{Hyung-Sik Kim}, \bibinfo{person}{Sang-Pyo Hong},
  \bibinfo{person}{Na-Rae Yu}, \bibinfo{person}{Dae-Woon Lim},
  \bibinfo{person}{Byung-Chan Min}, \bibinfo{person}{Gye-Rae Tack},
  {et~al\mbox{.}}} \bibinfo{year}{2013}\natexlab{}.
\newblock \showarticletitle{The effects of disruption in attention on driving
  performance patterns: Analysis of jerk-cost function and vehicle control
  data}.
\newblock \bibinfo{journal}{\emph{Applied ergonomics}} \bibinfo{volume}{44},
  \bibinfo{number}{4} (\bibinfo{year}{2013}), \bibinfo{pages}{538--543}.
\newblock


\bibitem[\protect\citeauthoryear{Dau and Keogh}{Dau and Keogh}{2017}]%
        {dau2017matrix}
\bibfield{author}{\bibinfo{person}{Hoang~Anh Dau} {and} \bibinfo{person}{Eamonn
  Keogh}.} \bibinfo{year}{2017}\natexlab{}.
\newblock \showarticletitle{Matrix profile V: A generic technique to
  incorporate domain knowledge into motif discovery}. In
  \bibinfo{booktitle}{\emph{Proceedings of the 23rd ACM SIGKDD International
  Conference on Knowledge Discovery and Data Mining}}.
  \bibinfo{pages}{125--134}.
\newblock


\bibitem[\protect\citeauthoryear{Fuermaier, Tucha, Evans, Koerts, de~Waard,
  Brookhuis, Aschenbrenner, Thome, Lange, and Tucha}{Fuermaier
  et~al\mbox{.}}{2017}]%
        {fuermaier2017driving}
\bibfield{author}{\bibinfo{person}{Anselm~BM Fuermaier}, \bibinfo{person}{Lara
  Tucha}, \bibinfo{person}{Ben~Lewis Evans}, \bibinfo{person}{Janneke Koerts},
  \bibinfo{person}{Dick de Waard}, \bibinfo{person}{Karel Brookhuis},
  \bibinfo{person}{Steffen Aschenbrenner}, \bibinfo{person}{Johannes Thome},
  \bibinfo{person}{Klaus~W Lange}, {and} \bibinfo{person}{Oliver Tucha}.}
  \bibinfo{year}{2017}\natexlab{}.
\newblock \showarticletitle{Driving and attention deficit hyperactivity
  disorder}.
\newblock \bibinfo{journal}{\emph{Journal of neural transmission}}
  \bibinfo{volume}{124}, \bibinfo{number}{1} (\bibinfo{year}{2017}),
  \bibinfo{pages}{55--67}.
\newblock


\bibitem[\protect\citeauthoryear{Grethlein and Onta{\~n}{\'o}n}{Grethlein and
  Onta{\~n}{\'o}n}{2020}]%
        {grethlein2020spatially}
\bibfield{author}{\bibinfo{person}{David Grethlein} {and}
  \bibinfo{person}{Santiago Onta{\~n}{\'o}n}.} \bibinfo{year}{2020}\natexlab{}.
\newblock \showarticletitle{Spatially Aligned Clustering of Driving Simulator
  Data}. In \bibinfo{booktitle}{\emph{The Thirty-Third International Flairs
  Conference}}.
\newblock


\bibitem[\protect\citeauthoryear{Groom, Van~Loon, Daley, Chapman, and
  Hollis}{Groom et~al\mbox{.}}{2015}]%
        {groom2015driving}
\bibfield{author}{\bibinfo{person}{Madeleine~J Groom}, \bibinfo{person}{Editha
  Van~Loon}, \bibinfo{person}{David Daley}, \bibinfo{person}{Peter Chapman},
  {and} \bibinfo{person}{Chris Hollis}.} \bibinfo{year}{2015}\natexlab{}.
\newblock \showarticletitle{Driving behaviour in adults with attention
  deficit/hyperactivity disorder}.
\newblock \bibinfo{journal}{\emph{BMC psychiatry}} \bibinfo{volume}{15},
  \bibinfo{number}{1} (\bibinfo{year}{2015}), \bibinfo{pages}{175}.
\newblock


\bibitem[\protect\citeauthoryear{Gwak, Shino, and Hirao}{Gwak
  et~al\mbox{.}}{2018}]%
        {gwak2018early}
\bibfield{author}{\bibinfo{person}{Jongseong Gwak}, \bibinfo{person}{Motoki
  Shino}, {and} \bibinfo{person}{Akinari Hirao}.}
  \bibinfo{year}{2018}\natexlab{}.
\newblock \showarticletitle{Early detection of driver drowsiness utilizing
  machine learning based on physiological signals, behavioral measures, and
  driving performance}. In \bibinfo{booktitle}{\emph{2018 21st International
  Conference on Intelligent Transportation Systems (ITSC)}}. IEEE,
  \bibinfo{pages}{1794--1800}.
\newblock


\bibitem[\protect\citeauthoryear{Hori, Watanabe, Hori, Harsham, Hershey, Koji,
  Fujii, and Furumoto}{Hori et~al\mbox{.}}{2016}]%
        {hori2016driver}
\bibfield{author}{\bibinfo{person}{Chiori Hori}, \bibinfo{person}{Shinji
  Watanabe}, \bibinfo{person}{Takaaki Hori}, \bibinfo{person}{Bret~A Harsham},
  \bibinfo{person}{JohnR Hershey}, \bibinfo{person}{Yusuke Koji},
  \bibinfo{person}{Yoichi Fujii}, {and} \bibinfo{person}{Yuki Furumoto}.}
  \bibinfo{year}{2016}\natexlab{}.
\newblock \showarticletitle{Driver confusion status detection using recurrent
  neural networks}. In \bibinfo{booktitle}{\emph{2016 IEEE International
  Conference on Multimedia and Expo (ICME)}}. IEEE, \bibinfo{pages}{1--6}.
\newblock


\bibitem[\protect\citeauthoryear{Iannaccone, Hauser, Ball, Brandeis, Walitza,
  and Brem}{Iannaccone et~al\mbox{.}}{2015}]%
        {iannaccone2015classifying}
\bibfield{author}{\bibinfo{person}{Reto Iannaccone}, \bibinfo{person}{Tobias~U
  Hauser}, \bibinfo{person}{Juliane Ball}, \bibinfo{person}{Daniel Brandeis},
  \bibinfo{person}{Susanne Walitza}, {and} \bibinfo{person}{Silvia Brem}.}
  \bibinfo{year}{2015}\natexlab{}.
\newblock \showarticletitle{Classifying adolescent
  attention-deficit/hyperactivity disorder (ADHD) based on functional and
  structural imaging}.
\newblock \bibinfo{journal}{\emph{European child \& adolescent psychiatry}}
  \bibinfo{volume}{24}, \bibinfo{number}{10} (\bibinfo{year}{2015}),
  \bibinfo{pages}{1279--1289}.
\newblock


\bibitem[\protect\citeauthoryear{IIHS}{IIHS}{2019}]%
        {iihs2019}
\bibfield{author}{\bibinfo{person}{IIHS}.} \bibinfo{year}{2019}\natexlab{}.
\newblock \showarticletitle{Fatality Facts 2018: Teenagers}.
\newblock \bibinfo{journal}{\emph{Insurance Institute for Highway Safety}}
  (\bibinfo{date}{Dec} \bibinfo{year}{2019}).
\newblock
\urldef\tempurl%
\url{https://www.iihs.org/topics/fatality-statistics/detail/teenagers}
\showURL{%
\tempurl}


\bibitem[\protect\citeauthoryear{Keogh, Chakrabarti, Pazzani, and
  Mehrotra}{Keogh et~al\mbox{.}}{2001}]%
        {keogh2001dimensionality}
\bibfield{author}{\bibinfo{person}{Eamonn Keogh}, \bibinfo{person}{Kaushik
  Chakrabarti}, \bibinfo{person}{Michael Pazzani}, {and}
  \bibinfo{person}{Sharad Mehrotra}.} \bibinfo{year}{2001}\natexlab{}.
\newblock \showarticletitle{Dimensionality Reduction for Fast Similarity Search
  in Large Time Series Databases}.
\newblock \bibinfo{journal}{\emph{Knowledge and Information Systems Journal
  (KAIS)}}.
\newblock
\urldef\tempurl%
\url{https://www.microsoft.com/en-us/research/publication/dimensionality-reduction-for-fast-similarity-search-in-large-time-series-databases/}
\showURL{%
\tempurl}


\bibitem[\protect\citeauthoryear{Kochanek, Murphy, Xu, and Arias}{Kochanek
  et~al\mbox{.}}{2019}]%
        {kochanek2019national}
\bibfield{author}{\bibinfo{person}{KD Kochanek}, \bibinfo{person}{SL Murphy},
  \bibinfo{person}{J Xu}, {and} \bibinfo{person}{E Arias}.}
  \bibinfo{year}{2019}\natexlab{}.
\newblock \bibinfo{title}{Deaths: Final Data for 2017 Volume 68 Number 9}.
\newblock
\newblock
\urldef\tempurl%
\url{https://www.cdc.gov/nchs/data/nvsr/nvsr68/nvsr68_09-508.pdf}
\showURL{%
\tempurl}


\bibitem[\protect\citeauthoryear{Lee, McIntosh, Winston, Power, Huang,
  Onta{\~n}{\'o}n, and Gonzalez}{Lee et~al\mbox{.}}{2018}]%
        {lee2018design}
\bibfield{author}{\bibinfo{person}{Yi-Ching Lee}, \bibinfo{person}{Chelsea~Ward
  McIntosh}, \bibinfo{person}{Flaura Winston}, \bibinfo{person}{Thomas Power},
  \bibinfo{person}{Patty Huang}, \bibinfo{person}{Santiago Onta{\~n}{\'o}n},
  {and} \bibinfo{person}{Avelino Gonzalez}.} \bibinfo{year}{2018}\natexlab{}.
\newblock \showarticletitle{Design of an experimental protocol to examine
  medication non-adherence among young drivers diagnosed with ADHD: A driving
  simulator study}.
\newblock \bibinfo{journal}{\emph{Contemporary clinical trials communications}}
   \bibinfo{volume}{11} (\bibinfo{year}{2018}), \bibinfo{pages}{149--155}.
\newblock


\bibitem[\protect\citeauthoryear{Lin, Keogh, Lonardi, and Patel}{Lin
  et~al\mbox{.}}{2002}]%
        {lin2002finding}
\bibfield{author}{\bibinfo{person}{Jessica Lin}, \bibinfo{person}{Eamonn
  Keogh}, \bibinfo{person}{Stefano Lonardi}, {and} \bibinfo{person}{Pranav
  Patel}.} \bibinfo{year}{2002}\natexlab{}.
\newblock \showarticletitle{Finding motifs in time series}. In
  \bibinfo{booktitle}{\emph{Proc. of the 2nd Workshop on Temporal Data
  Mining}}. \bibinfo{pages}{53--68}.
\newblock


\bibitem[\protect\citeauthoryear{Liu, Yamada, Collier, and Sugiyama}{Liu
  et~al\mbox{.}}{2013}]%
        {liu2013change}
\bibfield{author}{\bibinfo{person}{Song Liu}, \bibinfo{person}{Makoto Yamada},
  \bibinfo{person}{Nigel Collier}, {and} \bibinfo{person}{Masashi Sugiyama}.}
  \bibinfo{year}{2013}\natexlab{}.
\newblock \showarticletitle{Change-point detection in time-series data by
  relative density-ratio estimation}.
\newblock \bibinfo{journal}{\emph{Neural Networks}}  \bibinfo{volume}{43}
  (\bibinfo{year}{2013}), \bibinfo{pages}{72--83}.
\newblock


\bibitem[\protect\citeauthoryear{Lohrer and Lienkamp}{Lohrer and
  Lienkamp}{2015}]%
        {lohrer2015building}
\bibfield{author}{\bibinfo{person}{Jurgen Lohrer} {and} \bibinfo{person}{Markus
  Lienkamp}.} \bibinfo{year}{2015}\natexlab{}.
\newblock \showarticletitle{Building representative velocity profiles using
  FastDTW and spectral clustering}. In \bibinfo{booktitle}{\emph{2015 14th
  International Conference on ITS Telecommunications (ITST)}}. IEEE,
  \bibinfo{pages}{45--49}.
\newblock


\bibitem[\protect\citeauthoryear{Manawadu, Kawano, Murata, Kamezaki, Muramatsu,
  and Sugano}{Manawadu et~al\mbox{.}}{2018}]%
        {manawadu2018multiclass}
\bibfield{author}{\bibinfo{person}{Udara~E Manawadu}, \bibinfo{person}{Takahiro
  Kawano}, \bibinfo{person}{Shingo Murata}, \bibinfo{person}{Mitsuhiro
  Kamezaki}, \bibinfo{person}{Junya Muramatsu}, {and} \bibinfo{person}{Shigeki
  Sugano}.} \bibinfo{year}{2018}\natexlab{}.
\newblock \showarticletitle{Multiclass classification of driver perceived
  workload using long short-term memory based recurrent neural network}. In
  \bibinfo{booktitle}{\emph{2018 IEEE Intelligent Vehicles Symposium (IV)}}.
  IEEE, \bibinfo{pages}{1--6}.
\newblock


\bibitem[\protect\citeauthoryear{McDonald, Curry, Kandadai, Sommers, and
  Winston}{McDonald et~al\mbox{.}}{2014}]%
        {mcdonald2014comparison}
\bibfield{author}{\bibinfo{person}{Catherine~C McDonald},
  \bibinfo{person}{Allison~E Curry}, \bibinfo{person}{Venk Kandadai},
  \bibinfo{person}{Marilyn~S Sommers}, {and} \bibinfo{person}{Flaura~K
  Winston}.} \bibinfo{year}{2014}\natexlab{}.
\newblock \showarticletitle{Comparison of teen and adult driver crash scenarios
  in a nationally representative sample of serious crashes}.
\newblock \bibinfo{journal}{\emph{Accident Analysis \& Prevention}}
  \bibinfo{volume}{72} (\bibinfo{year}{2014}), \bibinfo{pages}{302--308}.
\newblock


\bibitem[\protect\citeauthoryear{McDonald, Tanenbaum, Lee, Fisher, Mayhew, and
  Winston}{McDonald et~al\mbox{.}}{2012}]%
        {mcdonald2012using}
\bibfield{author}{\bibinfo{person}{Catherine~C McDonald},
  \bibinfo{person}{Jason~B Tanenbaum}, \bibinfo{person}{Yi-Ching Lee},
  \bibinfo{person}{Donald~L Fisher}, \bibinfo{person}{Daniel~R Mayhew}, {and}
  \bibinfo{person}{Flaura~K Winston}.} \bibinfo{year}{2012}\natexlab{}.
\newblock \showarticletitle{Using crash data to develop simulator scenarios for
  assessing novice driver performance}.
\newblock \bibinfo{journal}{\emph{Transportation research record}}
  \bibinfo{volume}{2321}, \bibinfo{number}{1} (\bibinfo{year}{2012}),
  \bibinfo{pages}{73--78}.
\newblock


\bibitem[\protect\citeauthoryear{Mueen, Keogh, and Young}{Mueen
  et~al\mbox{.}}{2011}]%
        {mueen2011logical}
\bibfield{author}{\bibinfo{person}{Abdullah Mueen}, \bibinfo{person}{Eamonn
  Keogh}, {and} \bibinfo{person}{Neal Young}.} \bibinfo{year}{2011}\natexlab{}.
\newblock \showarticletitle{Logical-shapelets: an expressive primitive for time
  series classification}. In \bibinfo{booktitle}{\emph{Proceedings of the 17th
  ACM SIGKDD international conference on Knowledge discovery and data mining}}.
  \bibinfo{pages}{1154--1162}.
\newblock


\bibitem[\protect\citeauthoryear{NCSA}{NCSA}{2019}]%
        {US_DOT_19}
\bibfield{author}{\bibinfo{person}{NCSA}.} \bibinfo{year}{2019}\natexlab{}.
\newblock \bibinfo{title}{2018 Fatal Motor Vehicle Crashes: Overview, Traffic
  Safety Facts Research Note. Report No. DOT HS 812 826}.
\newblock
\newblock
\urldef\tempurl%
\url{https://crashstats.nhtsa.dot.gov/Api/Public/ViewPublication/812826}
\showURL{%
\tempurl}


\bibitem[\protect\citeauthoryear{Ontan{\'o}n, Lee, Snodgrass, Winston, and
  Gonzalez}{Ontan{\'o}n et~al\mbox{.}}{2017}]%
        {ontanon2017learning}
\bibfield{author}{\bibinfo{person}{Santiago Ontan{\'o}n},
  \bibinfo{person}{Yi-Ching Lee}, \bibinfo{person}{Sam Snodgrass},
  \bibinfo{person}{Flaura~K Winston}, {and} \bibinfo{person}{Avelino~J
  Gonzalez}.} \bibinfo{year}{2017}\natexlab{}.
\newblock \showarticletitle{Learning to predict driver behavior from
  observation}. In \bibinfo{booktitle}{\emph{2017 AAAI Spring Symposium
  Series}}.
\newblock


\bibitem[\protect\citeauthoryear{RaviPrakash, Watane, Jambawalikar, and
  Bagci}{RaviPrakash et~al\mbox{.}}{2019}]%
        {raviprakash2019deep}
\bibfield{author}{\bibinfo{person}{Harish RaviPrakash}, \bibinfo{person}{Arjun
  Watane}, \bibinfo{person}{Sachin Jambawalikar}, {and} \bibinfo{person}{Ulas
  Bagci}.} \bibinfo{year}{2019}\natexlab{}.
\newblock \showarticletitle{Deep Learning for Functional Brain Connectivity:
  Are We There Yet?}
\newblock In \bibinfo{booktitle}{\emph{Deep Learning and Convolutional Neural
  Networks for Medical Imaging and Clinical Informatics}}.
  \bibinfo{publisher}{Springer}, \bibinfo{pages}{347--365}.
\newblock


\bibitem[\protect\citeauthoryear{Sakoe and Chiba}{Sakoe and Chiba}{1974}]%
        {sakoe1974computer}
\bibfield{author}{\bibinfo{person}{Hiroaki Sakoe} {and} \bibinfo{person}{Seibi
  Chiba}.} \bibinfo{year}{1974}\natexlab{}.
\newblock \bibinfo{title}{Computer for calculating the similarity between
  patterns and pattern recognition system comprising the similarity computer}.
\newblock
\newblock
\newblock
\shownote{US Patent 3,816,722.}


\bibitem[\protect\citeauthoryear{Salvador and Chan}{Salvador and Chan}{2007}]%
        {salvador2007toward}
\bibfield{author}{\bibinfo{person}{Stan Salvador} {and} \bibinfo{person}{Philip
  Chan}.} \bibinfo{year}{2007}\natexlab{}.
\newblock \showarticletitle{Toward accurate dynamic time warping in linear time
  and space}.
\newblock \bibinfo{journal}{\emph{Intelligent Data Analysis}}
  \bibinfo{volume}{11}, \bibinfo{number}{5} (\bibinfo{year}{2007}),
  \bibinfo{pages}{561--580}.
\newblock


\bibitem[\protect\citeauthoryear{Sobanski, Sabljic, Alm, Skopp, Kettler,
  Mattern, and Strohbeck-K{\"u}hner}{Sobanski et~al\mbox{.}}{2008}]%
        {sobanski2008driving}
\bibfield{author}{\bibinfo{person}{E Sobanski}, \bibinfo{person}{D Sabljic},
  \bibinfo{person}{B Alm}, \bibinfo{person}{Gisela Skopp}, \bibinfo{person}{N
  Kettler}, \bibinfo{person}{Rainer Mattern}, {and} \bibinfo{person}{Peter
  Strohbeck-K{\"u}hner}.} \bibinfo{year}{2008}\natexlab{}.
\newblock \showarticletitle{Driving-related risks and impact of methylphenidate
  treatment on driving in adults with attention-deficit/hyperactivity disorder
  (ADHD)}.
\newblock \bibinfo{journal}{\emph{Journal of neural transmission}}
  \bibinfo{volume}{115}, \bibinfo{number}{2} (\bibinfo{year}{2008}),
  \bibinfo{pages}{347--356}.
\newblock


\bibitem[\protect\citeauthoryear{Srivastava, Hinton, Krizhevsky, Sutskever, and
  Salakhutdinov}{Srivastava et~al\mbox{.}}{2014}]%
        {dropout}
\bibfield{author}{\bibinfo{person}{Nitish Srivastava},
  \bibinfo{person}{Geoffrey Hinton}, \bibinfo{person}{Alex Krizhevsky},
  \bibinfo{person}{Ilya Sutskever}, {and} \bibinfo{person}{Ruslan
  Salakhutdinov}.} \bibinfo{year}{2014}\natexlab{}.
\newblock \showarticletitle{Dropout: A Simple Way to Prevent Neural Networks
  from Overfitting}.
\newblock \bibinfo{journal}{\emph{Journal of Machine Learning Research}}
  \bibinfo{volume}{15}, \bibinfo{number}{56} (\bibinfo{year}{2014}),
  \bibinfo{pages}{1929--1958}.
\newblock
\urldef\tempurl%
\url{http://jmlr.org/papers/v15/srivastava14a.html}
\showURL{%
\tempurl}


\bibitem[\protect\citeauthoryear{Tenev, Markovska-Simoska, Kocarev,
  Pop-Jordanov, M{\"u}ller, and Candrian}{Tenev et~al\mbox{.}}{2014}]%
        {tenev2014machine}
\bibfield{author}{\bibinfo{person}{Aleksandar Tenev}, \bibinfo{person}{Silvana
  Markovska-Simoska}, \bibinfo{person}{Ljupco Kocarev}, \bibinfo{person}{Jordan
  Pop-Jordanov}, \bibinfo{person}{Andreas M{\"u}ller}, {and}
  \bibinfo{person}{Gian Candrian}.} \bibinfo{year}{2014}\natexlab{}.
\newblock \showarticletitle{Machine learning approach for classification of
  ADHD adults}.
\newblock \bibinfo{journal}{\emph{International Journal of Psychophysiology}}
  \bibinfo{volume}{93}, \bibinfo{number}{1} (\bibinfo{year}{2014}),
  \bibinfo{pages}{162--166}.
\newblock


\bibitem[\protect\citeauthoryear{Verster and Roth}{Verster and Roth}{2011}]%
        {verster2011standard}
\bibfield{author}{\bibinfo{person}{Joris~C Verster} {and}
  \bibinfo{person}{Thomas Roth}.} \bibinfo{year}{2011}\natexlab{}.
\newblock \showarticletitle{Standard operation procedures for conducting the
  on-the-road driving test, and measurement of the standard deviation of
  lateral position (SDLP)}.
\newblock \bibinfo{journal}{\emph{International journal of general medicine}}
  \bibinfo{volume}{4} (\bibinfo{year}{2011}), \bibinfo{pages}{359}.
\newblock


\bibitem[\protect\citeauthoryear{Walshe, Oppenheimer, Kandadai, and
  Winston}{Walshe et~al\mbox{.}}{2019a}]%
        {walshe2019comparison}
\bibfield{author}{\bibinfo{person}{Elizabeth Walshe}, \bibinfo{person}{Natalie
  Oppenheimer}, \bibinfo{person}{Venk Kandadai}, {and}
  \bibinfo{person}{Flaura~K Winston}.} \bibinfo{year}{2019}\natexlab{a}.
\newblock \showarticletitle{Comparison of virtual driving test performance and
  on-road examination for licensure performance: a replication study}.
\newblock  (\bibinfo{year}{2019}).
\newblock


\bibitem[\protect\citeauthoryear{Walshe, Ward~McIntosh, Romer, and
  Winston}{Walshe et~al\mbox{.}}{2017}]%
        {walshe2017executive}
\bibfield{author}{\bibinfo{person}{Elizabeth~A Walshe},
  \bibinfo{person}{Chelsea Ward~McIntosh}, \bibinfo{person}{Daniel Romer},
  {and} \bibinfo{person}{Flaura~K Winston}.} \bibinfo{year}{2017}\natexlab{}.
\newblock \showarticletitle{Executive function capacities, negative driving
  behavior and crashes in young drivers}.
\newblock \bibinfo{journal}{\emph{International journal of environmental
  research and public health}} \bibinfo{volume}{14}, \bibinfo{number}{11}
  (\bibinfo{year}{2017}), \bibinfo{pages}{1314}.
\newblock


\bibitem[\protect\citeauthoryear{Walshe, Winston, Betancourt, Khurana, Arena,
  and Romer}{Walshe et~al\mbox{.}}{2019b}]%
        {walshe2019working}
\bibfield{author}{\bibinfo{person}{Elizabeth~A Walshe},
  \bibinfo{person}{Flaura~K Winston}, \bibinfo{person}{Laura~M Betancourt},
  \bibinfo{person}{Atika Khurana}, \bibinfo{person}{Kristin Arena}, {and}
  \bibinfo{person}{Daniel Romer}.} \bibinfo{year}{2019}\natexlab{b}.
\newblock \showarticletitle{Working memory development and motor vehicle
  crashes in young drivers}.
\newblock \bibinfo{journal}{\emph{JAMA network open}} \bibinfo{volume}{2},
  \bibinfo{number}{9} (\bibinfo{year}{2019}),
  \bibinfo{pages}{e1911421--e1911421}.
\newblock


\bibitem[\protect\citeauthoryear{Wong, Hastings, Negy, Gonzalez,
  Onta{\~n}{\'o}n, and Lee}{Wong et~al\mbox{.}}{2018}]%
        {wong2018machine}
\bibfield{author}{\bibinfo{person}{Josiah Wong}, \bibinfo{person}{Lauren
  Hastings}, \bibinfo{person}{Kevin Negy}, \bibinfo{person}{Avelino~J
  Gonzalez}, \bibinfo{person}{Santiago Onta{\~n}{\'o}n}, {and}
  \bibinfo{person}{Yi-Ching Lee}.} \bibinfo{year}{2018}\natexlab{}.
\newblock \showarticletitle{Machine learning from observation to detect
  abnormal driving behavior in humans}. In \bibinfo{booktitle}{\emph{The
  Thirty-First International Flairs Conference}}.
\newblock


\bibitem[\protect\citeauthoryear{Wu and Keogh}{Wu and Keogh}{2020}]%
        {wu2020fastdtw}
\bibfield{author}{\bibinfo{person}{Renjie Wu} {and} \bibinfo{person}{Eamonn~J
  Keogh}.} \bibinfo{year}{2020}\natexlab{}.
\newblock \showarticletitle{FastDTW is approximate and Generally Slower than
  the Algorithm it Approximates}.
\newblock \bibinfo{journal}{\emph{arXiv preprint arXiv:2003.11246}}
  (\bibinfo{year}{2020}).
\newblock


\bibitem[\protect\citeauthoryear{Yankov, Keogh, Medina, Chiu, and
  Zordan}{Yankov et~al\mbox{.}}{2007}]%
        {yankov2007detecting}
\bibfield{author}{\bibinfo{person}{Dragomir Yankov}, \bibinfo{person}{Eamonn
  Keogh}, \bibinfo{person}{Jose Medina}, \bibinfo{person}{Bill Chiu}, {and}
  \bibinfo{person}{Victor Zordan}.} \bibinfo{year}{2007}\natexlab{}.
\newblock \showarticletitle{Detecting time series motifs under uniform
  scaling}. In \bibinfo{booktitle}{\emph{Proceedings of the 13th ACM SIGKDD
  international conference on Knowledge discovery and data mining}}.
  \bibinfo{pages}{844--853}.
\newblock


\bibitem[\protect\citeauthoryear{Ye and Keogh}{Ye and Keogh}{2009}]%
        {ye2009time}
\bibfield{author}{\bibinfo{person}{Lexiang Ye} {and} \bibinfo{person}{Eamonn
  Keogh}.} \bibinfo{year}{2009}\natexlab{}.
\newblock \showarticletitle{Time series shapelets: a new primitive for data
  mining}. In \bibinfo{booktitle}{\emph{Proceedings of the 15th ACM SIGKDD
  international conference on Knowledge discovery and data mining}}.
  \bibinfo{pages}{947--956}.
\newblock


\bibitem[\protect\citeauthoryear{Yeh, Zhu, Ulanova, Begum, Ding, Dau, Silva,
  Mueen, and Keogh}{Yeh et~al\mbox{.}}{2016}]%
        {yeh2016matrix}
\bibfield{author}{\bibinfo{person}{Chin-Chia~Michael Yeh}, \bibinfo{person}{Yan
  Zhu}, \bibinfo{person}{Liudmila Ulanova}, \bibinfo{person}{Nurjahan Begum},
  \bibinfo{person}{Yifei Ding}, \bibinfo{person}{Hoang~Anh Dau},
  \bibinfo{person}{Diego~Furtado Silva}, \bibinfo{person}{Abdullah Mueen},
  {and} \bibinfo{person}{Eamonn Keogh}.} \bibinfo{year}{2016}\natexlab{}.
\newblock \showarticletitle{Matrix profile I: all pairs similarity joins for
  time series: a unifying view that includes motifs, discords and shapelets}.
  In \bibinfo{booktitle}{\emph{2016 IEEE 16th international conference on data
  mining (ICDM)}}. Ieee, \bibinfo{pages}{1317--1322}.
\newblock


\end{thebibliography}

\newpage
\appendix
\section*{Appendix}

This document provides supplementary information for the paper ``Identifying On-road Scenarios Predictive of ADHD using Driving Simulator Time Series Data''. This document contains supporting text and figures to the main article: 

\begin{itemize}
    \item Our uncovered evidence to support the claim that \emph{FastDTW} is not necessarily faster than \emph{DTW}\cite{wu2020fastdtw}.
    \item Detailed results from the individual experiments conducted during grid search.
\end{itemize}

\section{Comparing FastDTW to other DTW's}

For all \emph{Dynamic Time Warping} (DTW) similarity-based ISR experiments conducted, we pre-computed similarity matrices to quantify how uniformly driver participants behaved in encapsulated sections of the route. Since the simulated routes driven were partitioned into hundreds of sub-sections, constructing similarity matrices from the corresponding time series sub-intervals prior to training was a non-negligible computational overhead for building ISR ensembles.   

 Computing any formulation of similarity used in our experiments between time series \emph{A} and \emph{B} of lengths $N$ and $M \leq N$ should incur a \emph{time to compute} (TTC) that is $\Omega (N)$. Brute force DTW searches through all potential warpings in an $\mathcal{O}(NM) = \mathcal{O}(N^2)$ operation to find the minimum sum of residual differences. \emph{Sakoe-Chiba DTW} (SC DTW) approximates the true minimum sum of residual differences in $\mathcal{O}(N)$. SC DTW restricts the warping window of time-steps using a global \emph{radius} constraint. This equates to a band of cells $2*radius$ in width along the DTW cost matrix diagonal defining the warping window.  

FastDTW also takes a radius parameter to define a warping window twice its width to approximate the true minimum sum of residuals expressed by DTW. What is different is that the cells FastDTW uses to define a warping window were selected by recursively analyzing PAA down-samplings of the time series data and isolating alignments suspected to be nearly minimizing for further consideration. The hidden cost of reducing series to lower resolutions, finding near-minimum warping paths, projecting the path back to full resolution, and then refining the approximate path has recently been the subject of intense scrutiny\cite{wu2020fastdtw}.

$$error(Sim(A,B)) = \frac{|DTW(A,B) - Sim(A,B)|}{DTW(A,B)}$$

In both methods of approximating DTW the radius parameter tends to control the quality of the approximation; smaller radii trade a larger search area in the DTW cost matrix for a quicker TTC. The sacrifice in precision when computing approximate DTW similarity $Sim(A,B)$ is measurable as \emph{error}. One of the purposes of performing similarity-based classification is to leverage differences in the classes of time series data directly instead of laboriously refining domain-specific features to form our comparisons. 

ISR ensembles built using many different similarity metrics were intended to answer the question: ``how precisely must we compare driving simulator time series data in order to pick out teenagers driving with untreated ADHD?'' Permuting the hyper-parameters in a grid search was designed to assert ISR's maximal ability to differentiate the classes of drivers in our dataset. 

\begin{figure*}[t]
    \centering
    \includegraphics[width=.6\linewidth]{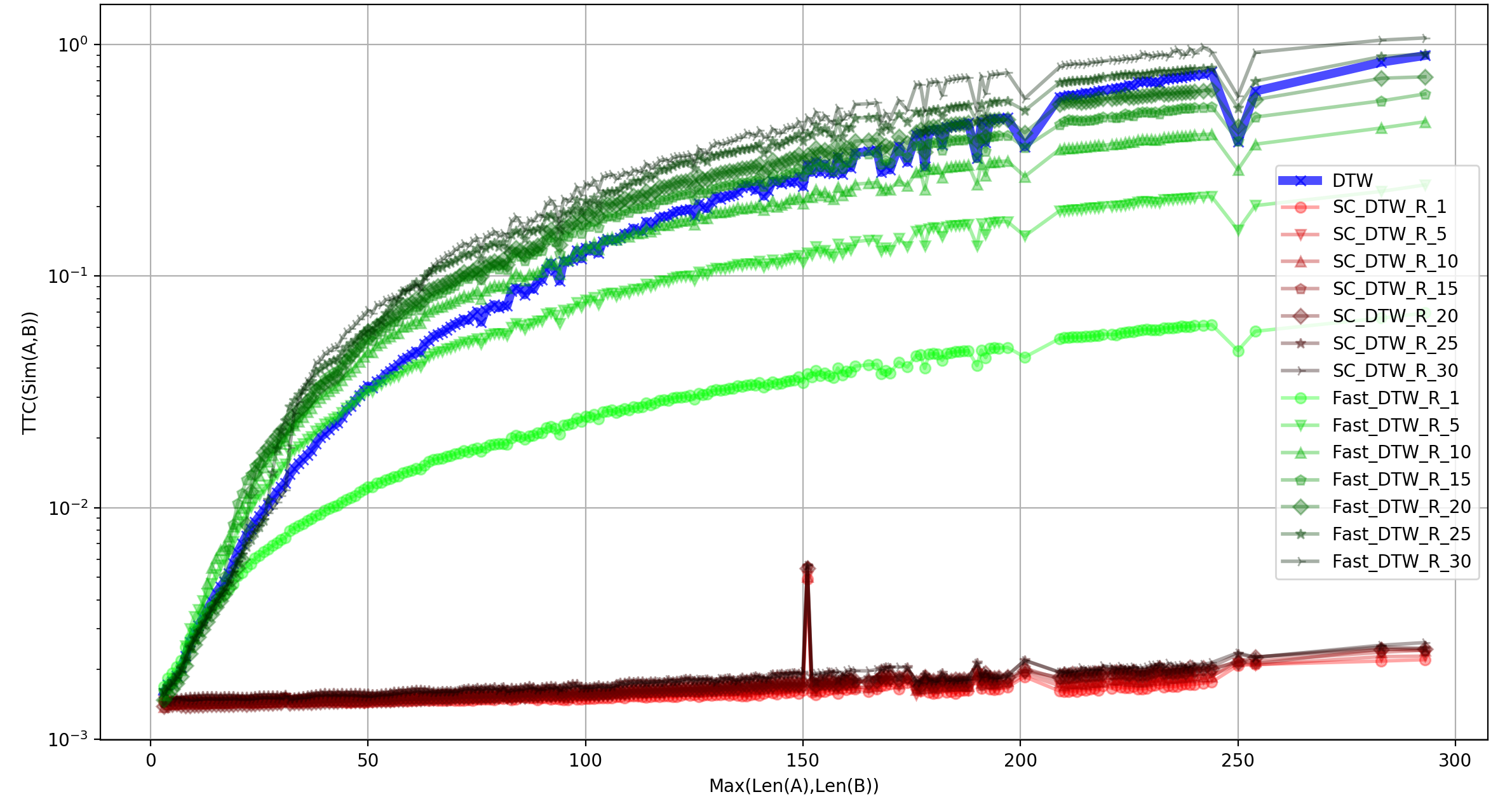}
    \caption{The average \emph{time to compute} (TTC) similarity in seconds (log scale) between two time series versus the length of the longer series in frames. All time series clippings were down-sampled via PAA to 1 Hz for the similarity-based experiments, so the number of frames in length correspond to the number of seconds in duration of recorded data.}
    \label{fig:ttc_plot}
\end{figure*}

\begin{figure*}[t]
    \centering
    \includegraphics[width=.6\linewidth]{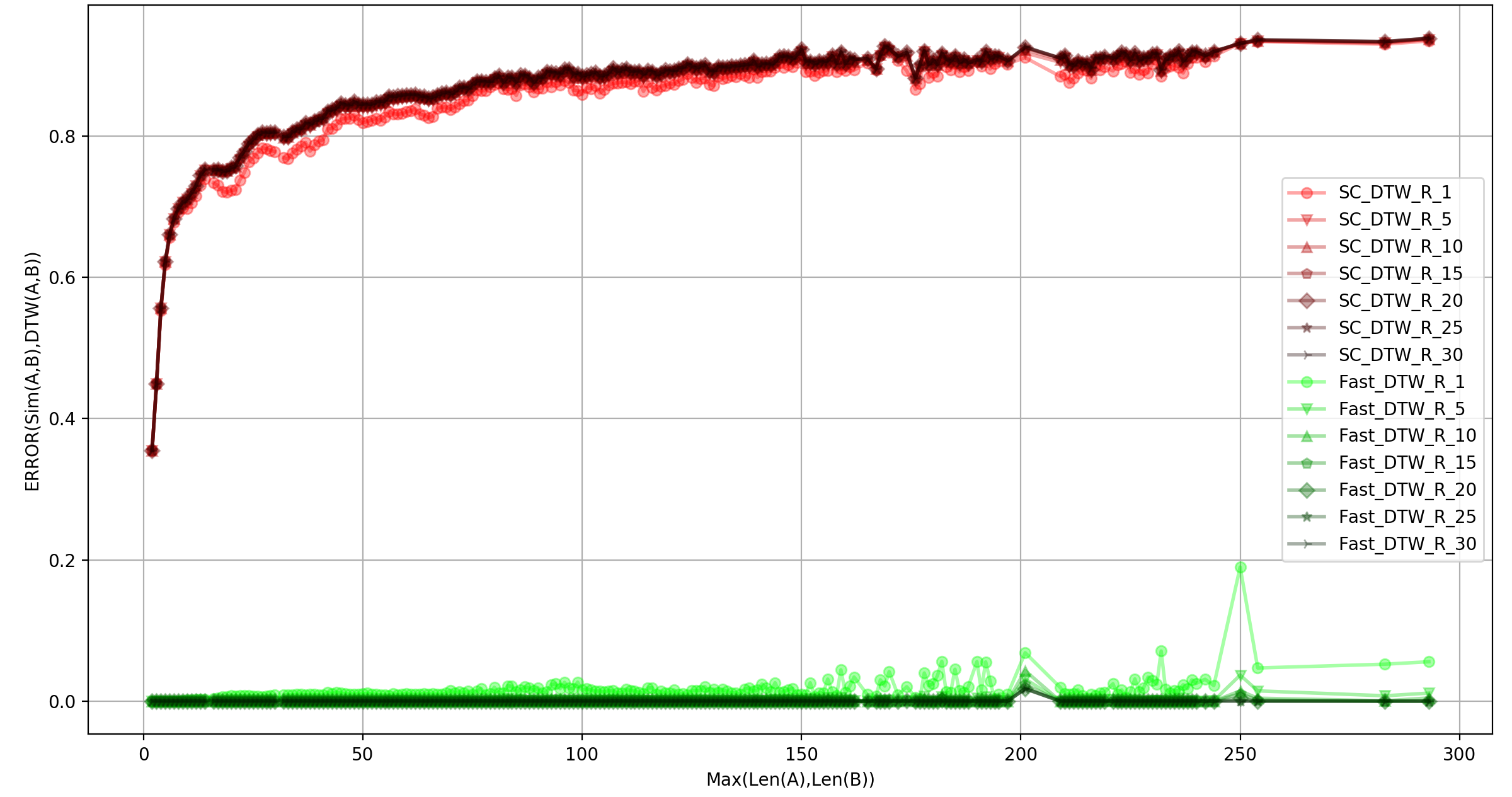}
    \caption{The average \emph{error} in each similarity metric's approximation to the true minimum sum of residuals computed by \emph{dynamic time warping} (DTW). All \emph{Sakoe-Chiba}  (SC DTW) comparisons produced significant error in approximating DTW. Nearly all FastDTW comparisons yielded a warping path that nearly matched the true minimum DTW sum of residuals.} 
    \label{fig:error_plot}
\end{figure*}

\subsection{Similarity-Based Experiments}

It is important to clarify that what we report as TTC is not the ``training time'' for any classifiers, though TTC was certainly a contributor to the overall time spent forming ISR similarity-based ensembles. Nor is the error reported in any way an indication of ISR classification results, such results are located in the main article.

For each calculation of time series similarity we recorded the lengths of the two series being compared as well as the amount of time that similarity computation took to complete (TTC). Since we had also computed every such pair-wise comparison of time series sub-intervals using DTW, we were able to compute the error generated in both the SC DTW and FastDTW approximations to it.

From the 31 top-level sections we built similarity matrices using up to 4 levels of ISR tree sub-sections for 15 different similarity metrics. Stored in the resulting 18,600 constructed similarity matrices are roughly 19 million pair-wise comparisons of time series. In order to characterize the trade-off between TTC and error when computing time series similarity, both were plotted against the maximum length of the series being compared. Each plotted data-point represents an average over all pair-wise comparisons for a particular similarity metric with a given max time series length.  

\subsection{Reproducibility and Implementation Details}

All experiments were ran on a central lab server using \emph{Intel Xeon E5-2699 v4} processors with 80 total cores clocked at 3.6 GHz, and 32 GB of RAM. These computing specs were far from fully utilized in building ensembles and evaluating them; experiments were limited to tapping into a maximum of 10 cores at a time. We ran our experiments in a \emph{Python 3.6.9} virtual environment, using similarity metric implementations from the \emph{fastdtw} and \emph{tslearn} modules. \emph{DeepLSTM} modules were constructed in the \emph{Keras} framework.

\subsection{FastDTW Can Be Slower than DTW}

Overall SC DTW, with the radii we tested, took at least one order of magnitude less time to compute similarity between time series clippings than both DTW and FastDTW. Our results for SC DTW were consistent with our understanding that TTC grows linearly with respect to the max length of the time series being compared (see Figure \ref{fig:ttc_plot}).  The small values assigned to the SC DTW radius parameter defined a narrow band of potential cell alignments. For longer time series we anticipated and observed a significant reduction in TTC with respect to DTW. While we also anticipated SC DTW would reduce TTC compared to FastDTW similarity, it was unclear whether the two values would grow disparately apart for longer series or would remain in step with one another. 

Figure \ref{fig:ttc_plot} shows that the TTC from computing FastDTW appears to grow at relatively the same rate as DTW, with certain parameterizations taking longer than DTW. All parameterizations of SC DTW produce TTC that grows linearly with the max length of the series compared (flattened on a log scale). As the radius parameter value was increased, both SC DTW and FastDTW's overall TTC increased as the warping window grew in size to cover a larger fraction of the DTW cost matrix. We did not test if these results hold for larger radius parameter values.

\subsection{Does FastDTW Closely Approximate DTW?}

One of the supposed benefits of using SC DTW or FastDTW in place of DTW is that it yields a reasonably faithful approximation to the true minimum sum of residual differences. Our experiments were designed to to gauge how closely FastDTW and SC DTW would approximate the true DTW minimum sum of residuals measured as error. We anticipated that SC DTW would be a less precise approximation to DTW than FastDTW as a fewer total number of potential frame alignments are considered using SC DTW. 

Figure \ref{fig:error_plot} shows the average error incurred from all computations of similarity using SC DTW and FastDTW with respect to the length of the longer time series being compared. The majority of SC DTW similarity approximations diverged significantly ($error > 70\%$) from DTW. Shorter time series sub-intervals compared using SC DTW and radii nearly the same size as the lengths of the series still yielded substantial divergence from DTW ($35\% \leq error \leq 70\%$), though were much closer to the true minimum sum of residuals. In those cases the majority of cells in the DTW cost matrix were evaluated, increasing the likelihood of the warping window containing the true DTW warping path that optimizes the alignment of frames.

All FastDTW parameterizations resulted in very faithful ($error < 20\%$) approximations to the true DTW minimum sum of residuals. The smaller FastDTW radius values were the only ones to yield similarity with noticeable divergence from DTW ($error \geq 5\%$) when comparing longer time series. The FastDTW similarity-based ensembles performed marginally better than the SC DTW ensembles in classifying teen drivers with untreated ADHD from the simulator time series data. While our evidence is only statistically robust for one small dataset, it suggests that using less precise approximations of DTW similarity will form less accurate classifiers.

\end{document}